\documentclass[]{elsarticle}
\usepackage{hyperref}
\usepackage{amssymb}
\usepackage{graphicx}
\usepackage{subfig}
\usepackage{commath}
\usepackage{wrapfig}

\journal{Journal of Compuational Design and Engineering}

\begin{document}

\begin{frontmatter}

\title{Automatic Segmentation  of Spine using Convolutional Neural Networks via Redundant Generation of Class Labels}


\author[mymainaddress,mysecondaryaddress]{Malinda Vania \fnref{equalauthorship}}
\fntext[equalauthorship]{Both authors contributed equally to this work.}
\author[mythirdaddress]{Dawit Mureja \fnref{equalauthorship}}

\author[mymainaddress,mysecondaryaddress]{Deukhee Lee\corref{mycorrespondingauthor}}
\cortext[mycorrespondingauthor]{Corresponding author}
\ead{dyklee@kist.re.kr}

\address[mymainaddress]{Center for Bionics, Korea Institute of Science and Technology, Seoul 02792, Republic of Korea}
\address[mysecondaryaddress]{Division of Bio-Medical Science \& Technology, KIST School, Korea University of Science and Technology, Seoul 02792, Republic of Korea}
\address[mythirdaddress]{Korea Advanced Institute of Science and Technology, Daejeon 34141, Republic of Korea}

\begin{abstract}
There has been a significant increase from 2010 to 2016 in the number of people suffering from spine problems. The automatic image segmentation of the spine obtained from a computed tomography (CT) image is important for diagnosing spine conditions and for performing surgery with computer-assisted surgery systems. The spine has a complex anatomy that consists of 33 vertebrae, 23 intervertebral disks, the spinal cord, and connecting ribs. As a result, the spinal surgeon is faced with the challenge of needing a robust algorithm to segment and create a model of the spine. In this study, we developed an automatic segmentation method to segment the spine, and we compared our segmentation results with reference segmentations obtained by experts. We developed a fully automatic approach for spine segmentation from CT based on a hybrid method. This method combines the convolutional neural network (CNN) and fully convolutional network (FCN), and utilizes class redundancy as a soft constraint to greatly improve the segmentation results. The proposed method was found to significantly enhance the accuracy of the segmentation results and the system processing time. Our comparison was based on 12 measurements: the Dice coefficient (94\%), Jaccard index (93\%), volumetric similarity (96\%), sensitivity (97\%), specificity (99\%), precision (over segmentation; 8.3 and under segmentation 2.6), accuracy (99\%), Matthews correlation coefficient (0.93), mean surface distance (0.16 mm), Hausdorff distance (7.4 mm), and global consistency error (0.02). We experimented with CT images from 32 patients, and the experimental results demonstrated the efficiency of the proposed method.
\end{abstract}


\end{frontmatter}

\section{Introduction}
Computer-based technology plays an important role in defining how surgery is performed today \cite{ref3}. In computer-assisted surgery (CAS), the surgeon uses a surgical navigation system to navigate an instrument in relation to the anatomy of the patient. A computer-based system uses medical images, such as computed tomography (CT) images, of the patients to extract relevant information and create a 3D model of the patient. This model can be manipulated easily by the surgeon to provide views from any angle and at any depth within the volume. Thus, the surgeon can more thoroughly assess the situation and establish a more accurate diagnosis; such an approach is utilized in computer-assisted spinal diagnosis and therapy support systems \cite{ref2}.

The most important element of CAS is image segmentation. This process is used to construct an accurate model of the patient. Image segmentation is important for extracting information from an image. The segmentation process subdivides an image into its constituent parts or objects, depending on the problem to be solved. Segmentation is stopped when the region of interest in a specific application has been isolated.

One of the most difficult tasks in this process is the autonomous segmentation method. This step determines the eventual success or failure of the image analysis in which organ visualization is a critical aspect \cite{ref1,ref4}. Today, medical imaging modalities generate high resolutions and large number of images that cannot be examined manually. This drives the development of more efficient and robust problem-tailored image analysis methods for medical imaging. Automated image segmentation could increase precision by eliminating the subjectivity of the clinician. It also saves tremendous time and effort by eliminating an exhaustive process, where the results are hardly repeatable \cite{ref5}. 

The automatic spine segmentation process used to generate anatomically correct 3D models has challenges associated with it use. Some of these challenges are attributed to the anatomic complexity of the spine (33 vertebrae, 23 intervertebral disks, spinal cord, connecting ribs, etc.), image noise (all real-life data and CT images contain noise), low intensity (in spongy bones and softer bones), and the partial volume effect. 

Many methods have been proposed to alleviate these challenges in recent years \cite{ref1,ref6,ref8,ref11,ref12}. Recent spine segmentation research can be categorized into two main approaches: free estimation methods and trainable methods. Free-estimation methods do not require an explicit model for segmentations and include the following: classical region growing, watershed, active contours, and graph-cut methods \cite{ref1,ref8}. However, the trainable methods have a central assumption that the structures of interest/organs have repetitive geometry. Therefore, we can utilize the repetitive geometry into a probabilistic representation aimed toward explaining the variation in the shape of the organ and then when segmenting an image uses this information. 

Kang et al. \cite{ref13} utilized adaptive thresholding combined with region growing to conduct 3D bone segmentation. Mastmeyer et al. \cite{ref14} used a region growing method that was capable of detecting the disks between the vertebrae. Sambucetti et al. \cite{ref15} proposed a 2D active contour segmentation on a per-slice basis to construct 3D bone volume. All of these methods require expert human intervention and the manual adjustment of the parameter settings at several distinct steps. 

Several automatic methods have been proposed for vertebral column segmentation from CT images \cite{ref16,ref17,ref18,ref19,ref20}. Most of these methods consist of two steps: identification of the spine and separate individual segmentation of the spine vertebrae. Yao et al. \cite{ref16} used watershed segmentation and directed graph search methods to locate the vertebral body surfaces. The method was performed for several datasets and leakages occurred in 14 cases. Furthermore, identification and segmentation of the vertebrae were not carried out. Klinder et al. \cite{ref17} used a 3D deformable model through mesh adaptation. The disadvantages of this method lie in its dependency on tremendous parameter setup.
Recent advances in medical image segmentation techniques employ machine learning techniques to increase segmentation accuracy and gradually reduce human intervention. Huang et al. \cite{ref18} constructed vertebrae detectors by using Adaboost. Ma and Liu \cite{ref19} used low-level edge descriptors for vertebrae detection. Glocker et al. \cite{ref20} detected vertebrae shapes and labeled them using a model trained with supervised classification forest; however, this method required selecting an appropriate feature and relying on a priori knowledge of the spine shape. Therefore, this method is less applicable for general and varying image data. 

In this study, we propose the utilization of class redundancy combined with an improved hybrid of the convolutional neural network (CNN) and fully convolutional network (FCN) methods to overcome the drawbacks of previous methods and provide a practical solution. We present a fully automatic approach for spine segmentation from CT based on a hybrid method of CNN and FCN with the following main contributions:
\begin{enumerate}
	\item  We propose an efficient hybrid training scheme by utilizing a mask on sampled image segments and analyze its behavior in adapting to the class imbalance of the segmentation problem at hand.
	\item We demonstrate the capabilities of our system using class redundancy as soft constraints that greatly improves segmentation results. The efficiency of the proposed method will be demonstrated through the experimental results.
	
\end{enumerate}

This study is organized as follows. In section 2, we introduce histogram-based segmentation. Next, we explain the CNN and describe the proposed method in detail. Section 3 presents the experimental results and analysis. We evaluate the method on a database of 32 cases and compare the results with other automatic methods in order to draw reliable conclusions. Section 4 concludes the study and discusses results and future work.

\section{Methodology}

\subsection{Histogram-based level set segmentation}
Let the given CT data be represented as the following function, $f : \Omega \times \{1,.....,m\} \subset \mathbb{R}^{2}\times \mathbb{N}\rightarrow \mathbb{R}$, where $m$ is the number of sliced images in a volume data. Before segmenting, the CT data are preprocessed. This includes morphological image processing, region-based image processing, and contrast adjustment. After data preprocessing, histogram-based multiphase segmentation is performed. This method was first introduced by \cite{ref21} and has been used in many previous works \cite{ref22,ref23}.

The segmentation model in \cite{ref21} deals with a 1D histogram of the preprocessed data, and uses adaptive global maximum clustering (AGMC) to automatically obtain the number of significant local maxima of the histogram using 2-means clustering. This procedure provides n number of distinct regions and their corresponding subdomains. Then, a region-based segmentation is performed using a level-set function to label the different regions in the data. 

Once n-labeled data are obtained, the region of interest can be chosen differently by assigning a different label in accordance with the property of the desirable objects. Then, a detailed segmentation of the desired region is carried out using an active contour model, as suggested in \cite{ref22}. This model performs a segmentation based on a variable-weighted combination of local and global intensity. This will enable it to divide an object surrounded by both and weak boundaries and to distinguish very adjacent objects with those boundaries.

This method generally yields a good result, but it has several shortcomings. First, it is tedious. The original CT data have to pass through various segmentation phases in order to obtain the final desired result. Second, setting the optimal value of the parameters used in the segmentation models (\cite{ref21,ref22}) is not an easy task. Third, it is dependent on the specific CT dataset. Different spine CT datasets will have different numbers of distinct regions, based on the histogram-based multiphase segmentation. Hence, the manual selection of labels is necessary for different datasets. Some parameters are also different for different CT datasets. As a result, a segmentation model that would potentially eliminate the downsides of the current method is necessary.

\subsection{Convolutional neural networks}
A CNN \cite{ref28} is a type of feed-forward network that utilizes grid-like topology to analyze data. Through the use of local receptive fields, weight sharing and subsampling mechanisms, CNNs have proved themselves to be successful in various supervised tasks, such as image classification, object recognition \cite{ref29}, and image segmentation \cite{ref26,ref27}. In such tasks, models using CNNs are trained using an image dataset that is associated with a certain class label. In this regard, to perform spine segmentation using these networks, we first have to transform the CT data into an image dataset that can be analyzed by the networks. 

\subsubsection{Preparing training and testing data}
Because spine CT data are volumetric data, they are processed frame by frame. However, each frame of the CT data is not necessary for training the network, as the structure of the spine is symmetric across some interval of frames (the inter-vertebral disk). In order to prepare a label for the training data, we first segmented the training frames using the histogram-based level set segmentation method discussed in section 2.1. This can also be performed manually by using software programs such as 3D slicer. Once the ground truth was obtained, training data were prepared using 2D patching. Because the task involves only segmenting the spine that is located in a certain area of a frame, there is no need to process the entire frame. For computational simplicity, we formed a rectangular box around the area of the spine in each training frame. Note that the box size should be big enough to tolerate the spatial variation of the spine structure across the frames of the CT data. 

\begin{figure}
	\centering
	\begin{tabular}{cc}
		\subfloat[Training frame]{\includegraphics[scale = 0.32]{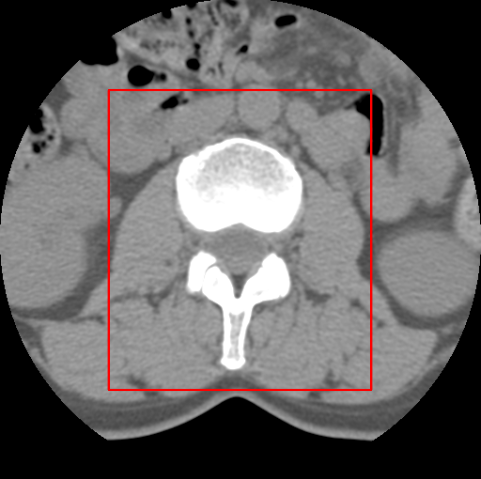}} &
		\subfloat[Ground truth]{\includegraphics[scale = 0.32]{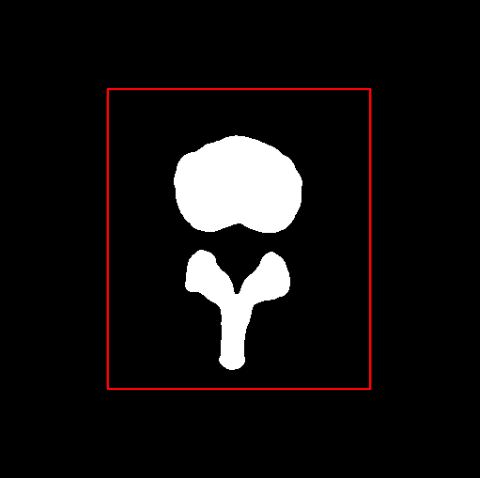}} 
			\end{tabular}

\caption{\label{fig:2dpatch} When preparing the dataset, we take a pixel inside the box of a given training frame and we will form a patch size of $n\times n$ around it. If the same pixel is part of the spine in the ground truth, then we label the patch as a spine (label 1). Otherwise, it will be labelled as background (label 0). We repeat the same procedure for all pixels in the box by taking a sliding interval of $s$. The test data is also prepared in a similar manner. This is called 2D patching.}
\end{figure}

\begin{wrapfigure}{L}{0.3\textwidth}
	\centering
	\includegraphics[width = 0.3\textwidth]{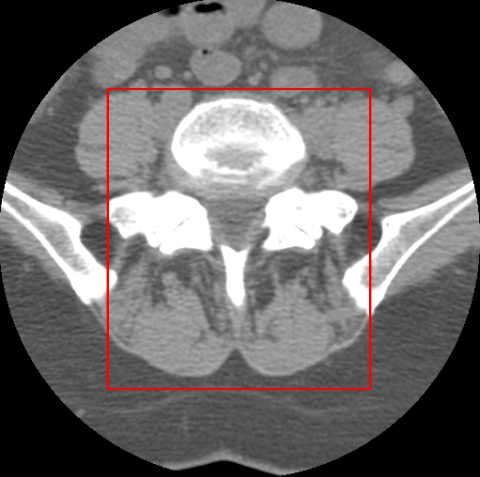}
	\caption{\label{fig:rib}A testing frame containing a rib.}
\end{wrapfigure} 
In 2D patching, we take a pixel inside the rectangular box of a given frame and form an $n\times n$ sized patch around it. By juxtaposition, if the same pixel in the ground truth is a part of the spine, then the patch will be labeled as 1 (class 1). Otherwise, it will be labeled as 0 (class 2). By taking a certain stride size of s, we repeated the same procedure across all pixels inside the box to prepare the training images (patches). Padding was not required to form the patches of the boundary pixels in the box because the respective distances between the edges of a given frame and the boundaries of the box were longer than n/2. Now, we have two classes of training data (0 and 1), and each class contains training patches obtained from the training frames. Testing data were also prepared in the same manner. We used the same rectangular box in the testing frames and then prepared the test images (patches) by using a 2D patching method. During testing, the model is expected to classify if a certain patch is part of the spine or not. Based on the result, we reconstruct the ground truth of each frame of the spine; in other words, we segment the spine (Fig. \ref{fig:2dpatch}). 

In the above scenario, the segmentation of the spine is treated as an image classification task with two classes. The results of this method are shown in the experimental section. Although the segmentation result was encouraging, it was not very accurate, particularly at the boundaries of the spine, as shown in the qualitative evaluation in section 3.3. In our work, we found out that using only two classes has certain downsides. First, there is an imbalance in the number of training images in the two classes prepared from the 2D patching. Because the spatial area of the spine in a given frame is small compared to the area of the background, most of the training patches are labeled as 0. Hence, the model eventually learns more about the background than the spine itself. The other problem is related to the pixel intensity. During testing, if the frame contains some parts of the rib (see Fig. \ref{fig:rib}), the model will most likely classify it as part of the spine (label 1) because the pixel intensity of the two is very similar in the CT data. Moreover, because we 
only have two classes, if the trained model is uncertain about the class a
certain patch belongs to, there is a 50\% chance it will commit an error.

\subsubsection{Proposed method: redundant generation of class labels}
\begin{figure}
	\centering
	\includegraphics[scale= 0.8]{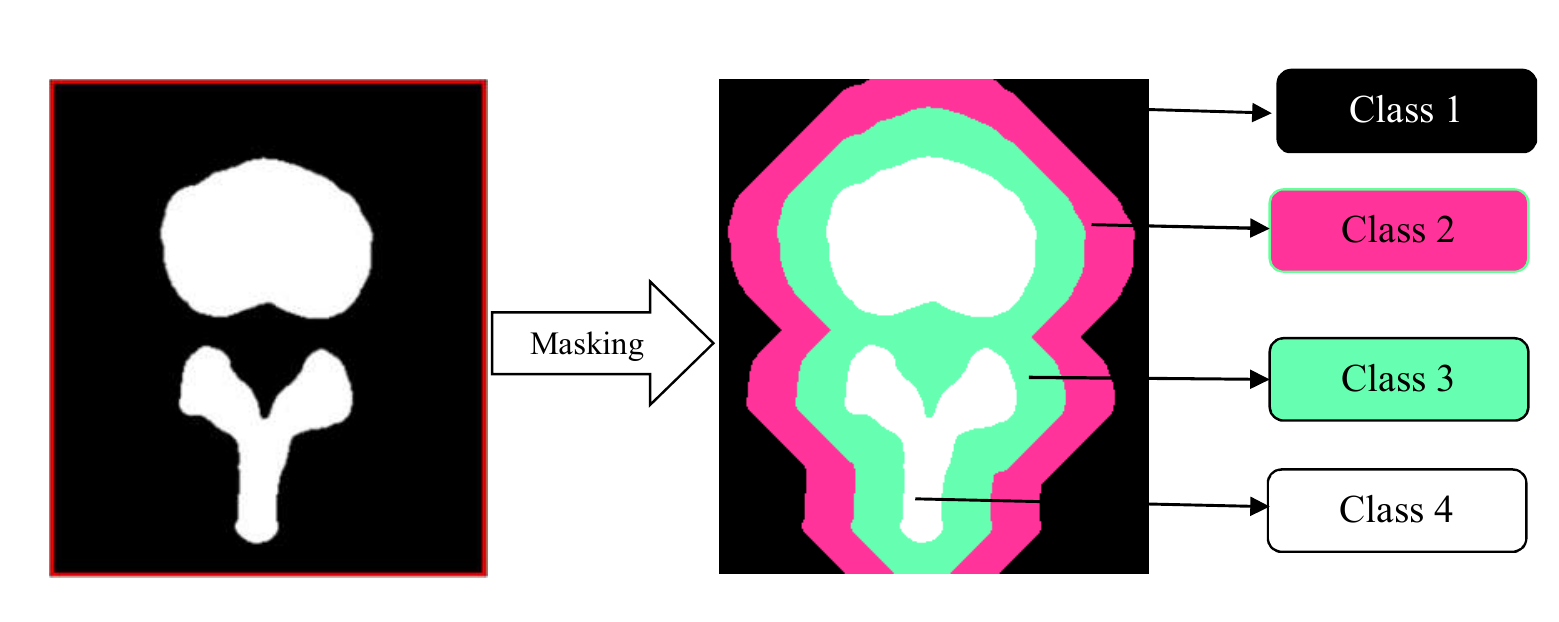}
	\caption{\label{fig:mask} Masking. We generate redundant classes by masking the spine in the ground truth of a given training frame. In this work, we used four classes for classification. Class 4 represents the spine. Class 1 represents the background. The other two classes are redundant. The first redundant class (class 3) was generated by masking the ground truth with a pixel value of 2. The second masking (class 2) was done with a pixel value of 3. The masking thickness was chosen so as to balance the data in each distinct class.}
\end{figure}

In order to address the above problems, we propose a new way of preparing the training data. The proposed method involves generating redundant class labels by masking the spine structure in each training frame. The ground truth of the training frames in Fig. \ref{fig:2dpatch} has a pixel value of 1 for the spine and a pixel value of 0 for the background. We generate the first redundant class by masking the area of the spine with a different pixel value (2, for instance). This class is important for training the model because it is used as a mechanism for punishing the model to accurately distinguish between the spine and its surrounding environment. In a similar manner, we generate more classes by continuing the masking with different pixel values, each associated with different class label (see Fig. \ref{fig:mask}). This enables us to obtain a proportional area of different classes in the ground truth. In general, the proposed method has the following advantages:  
\begin{itemize}
	\item A proportional amount of training patches can be prepared for each class.
	\item The model properly learns to segment the boundaries of the spine within a given frame because the masking is applied along the spine boundary.
	\item Anything inside the rectangular box that is not part of the spine will now have a specific label. For example, if the testing frame contains parts of the rib, the model will be able to identify those parts by the labels. This will avoid segmenting the rib along with the spine, if trained properly.
	\item The probability that the model makes errors will be reduced because we will have more classes.
	
\end{itemize}
\subsubsection{Model architecture}
\begin{figure*}
	\centering
	\includegraphics[scale= 0.5]{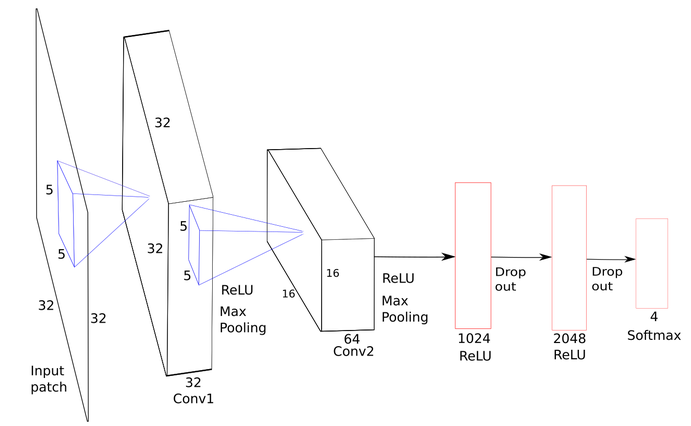}
	\caption{\label{fig:conv}Convolutional network architecture}
\end{figure*}

In this work, we used the architecture shown in Fig. \ref{fig:conv}. We implemented a simple neural architecture (2 CNN layers and 3 Fully Connected layers) compared to other deep networks used for medical image segmentation such as U-net (23 CNN layers) \cite{ref27}. Here is how our network works.
\begin{itemize}
	\item The input image (patch) has a size of $32\times 32$.  The first 2D convolution has a local receptive field size of $5\times 5$ and it outputs a convolutional feature map with 32 layers (neurons).  A zero-padding is used for boundary pixels while doing the convolution, hence, each layer in the feature map has the same size as the input image. A rectified linear unit (ReLU) activation layer that computes element wise non-linearity follows this layer. Later, a max pooling layer of size $2 \times 2$ subsamples the spatial size of the input patch. The output of this stage has a size of $16 \times 16 \times 32$.
	\item The second 2D convolution also has a local receptive field  size of $5\times 5$. However, it outputs a convolutional feature map of 64 layers (neurons). This layer is followed by a ReLU activation layer and a max pooling layer of size 2x2.  The output of this stage has a size of $8 \times 8 \times 64$.
	\item The output of the second stage is flattened and fed into a fully connected layer with 1024 neurons. This layer has a ReLU activation layer. Following this layer, a dropout is applied. Every time before the input is presented, the neural network drops out different set of neurons with probability of 0.5. This reduces the network's dependency on the presence of particular neurons.
	A second fully connected layer with 2048 neurons uses the output of the previous stage as an input. This layer also has a ReLU activation layer. A dropout is also implemented in a similar manner as the previous layer.
	
	\item The final layer is a softmax layer with 4 neurons. This layer outputs a probabilistic prediction array for the 4 classes we used. Note that, even though we treated spine segmentation as a 4-class classification problem, only one class is important for labeling the spine. The other three classes are irrelevant (redundant). 
\end{itemize}

\section{Experimental design and results}
There are several factors that affected the diversity of the medical image data, such as the imaging modalities, the type of CT-scan machine used, the radiation dose, the scan time, and the patients. In order to prove the robustness of our method, we tested our algorithm for a diverse array of patient datasets.

\subsection{Data sources}
We tested our method on several public datasets that were obtained from the Spineweb website and Gangnam Severance Hospital. The datasets were CT scan data with $512\times 512$ pixel images for 32 patients. To validate the segmentation result, we used the ground truth from Gangnam Severance Hospital. The dataset contained $512\times 512$ pixel CT images with 0.3125 mm thicknesses, which act as the gold standard in this project. The reason we used data from Spineweb is to ensure there is diversity in the real clinical data with respect to patients and the CT-Scan imaging machine.

\subsection{Parameter selection and training computational time}
The patch size was set to $32\times 32$ with a sliding interval of 1 pixel. Our network was implemented in Python and Tensorflow. All experiments were performed on a PC (GeForce GTX 1080 Ti. and 32 GB RAM). The training stage is time-consuming for CPU implementation. We compared our network with U-Net \cite{ref27}. U-Net took approximately 25 h for training and approximately 6.5 min (385 s) to process one image. However, these stages were completed faster in our network compared to U-Net where training the network took approximately 13 h and segmentation processing time took approximately 3 min (170 s) for each image. 

The training process was performed on patches that were obtained from the images (frames in the CT data). To obtain a large training dataset, patches were used instead of the entire image to train the network. In addition, patches can represent the local structures at a higher quality level. A corresponding label and 1.388.800 patches of spine data were selected randomly from the training dataset. To ensure fairness of the analysis results, the testing dataset that we used was different from the dataset that was used in the training set. The fairness in here is necessary in order to show that even if the system had never seen the input data, it can produce good segmentation results based on previous training data.

A number of measurement metrics used in the evaluation of segmentation were computed for each segmentation result \cite{ref24,ref25}. The metrics chosen for quantitative analysis were divided into similarity, classic, and distance measurements. The similarity metrics included the Dice coefficient (DC), Jaccard index, and volumetric similarity (VS). The classic measurements used sensitivity, specificity, precision (over segmentation (OS) and under segmentation (US)) and accuracy. The distance measurement used the mean surface distance (MSD), Hausdorff distance (HD), and the global consistency error (GCE). We also used the Matthews correlation coefficient (MCC).

In this work , we prepared the training and testing dataset using the method discussed in section 2.2.1. Then we used a simple CNN model discussed in section 2.2.2 to do the spine segmentation. We also used the prepared data with a rather deep neural network (U-net) \cite{ref27} for comparison purpose.  In addition, two different methods (level-set and CNN with two classes) were compared against our method.
\begin{figure}
	\centering
	\begin{tabular}{cccc}
		\subfloat[Our method]{\includegraphics[width = 1.125in, height= 1.575in]{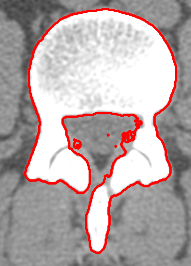}} &
		\subfloat[2-Class]{\includegraphics[width = 1.125in, height= 1.575in]{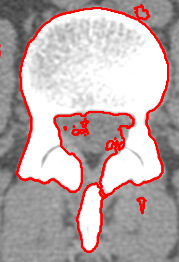}} &
		\subfloat[U-net]{\includegraphics[width = 1.125in, height= 1.575in]{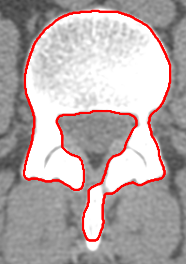}} &
		\subfloat[Level set]{\includegraphics[width = 1.125in, height= 1.575in]{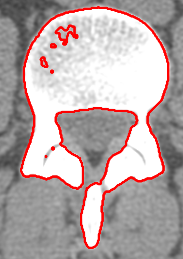}}\\
		\subfloat[Our method]{\includegraphics[width = 1.125in, height= 1.475in]{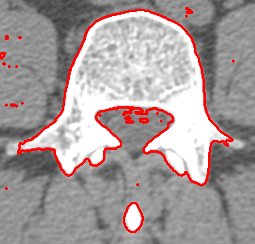}} &
		\subfloat[2-Class]{\includegraphics[width = 1.125in, height= 1.475in]{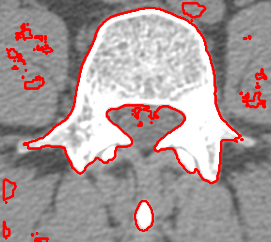}} &
		\subfloat[U-net]{\includegraphics[width = 1.125in, height= 1.475in]{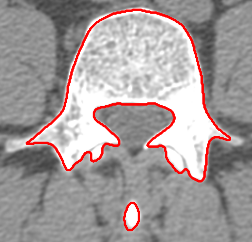}} &
		\subfloat[Level set]{\includegraphics[width = 1.125in, height= 1.475in]{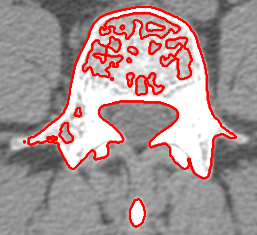}}\\
		\subfloat[Our method]{\includegraphics[width = 1.125in, height= 1.475in]{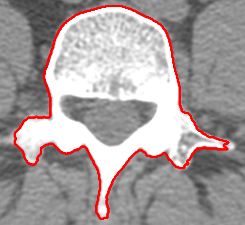}} &
		\subfloat[2-Class]{\includegraphics[width = 1.125in, height= 1.475in]{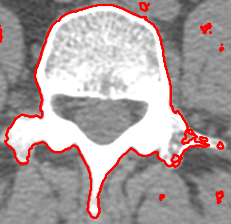}} &
		\subfloat[U-net]{\includegraphics[width = 1.125in, height= 1.475in]{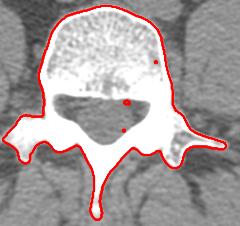}} &
		\subfloat[Level set]{\includegraphics[width = 1.125in, height= 1.475in]{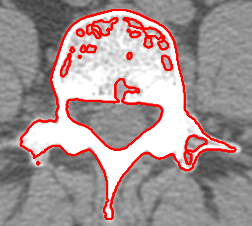}}\\
		\subfloat[Our method]{\includegraphics[width = 1.125in, height= 1.475in]{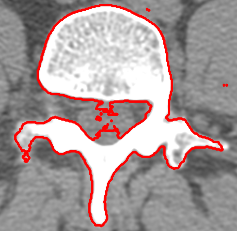}} &
		\subfloat[2-Class]{\includegraphics[width = 1.125in, height= 1.475in]{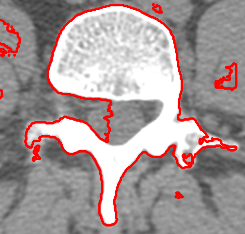}} &
		\subfloat[U-net]{\includegraphics[width = 1.125in, height= 1.475in]{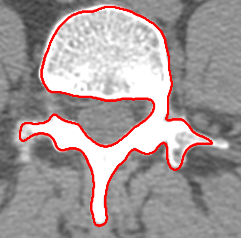}} &
		\subfloat[Level set]{\includegraphics[width = 1.125in, height= 1.475in]{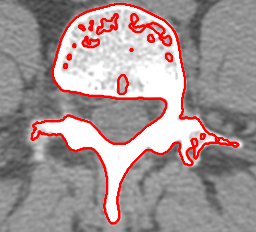}}\\
	\end{tabular}
	\vspace{-0.5cm}
	\caption{\label{fig:qualcomparison}Qualitative comparison between our method, 2-Class CNN, Level set and U-net model.}
\end{figure}
\subsection{Experimental results}
\subsubsection{Qualitative evaluation}
We selected a few representative slices from the results of the testing set. Fig. \ref{fig:qualcomparison} presents the results obtained using the different methods. The results demonstrate that the proposed method segments more accurately than the other methods.
\subsubsection{Quantitative evaluation}
The manually labeled images of each subject are used as gold standards and the results of each segmentation method are converted into binary images with the same voxel resolution and image dimensions as the query image. In the following description of the measures, the segmentation result is indicated by S and the gold standard by GT.
\begin{enumerate}[(A)]
	\item Similarity metrics
	\begin{enumerate}[(i)]
		\item Dice coefficient (DC)
		
	DC measures the extent of the spatial overlap between two binary images. DC values range between 0 (no overlap) and 1 (perfect agreement). In this study, the DC values are obtained using Equation (1).
		\begin{equation}
			DC = \frac{2\left|S\cap GT\right|}{\left|S\right| + \left|GT\right|}
		\end{equation}
	
	\item The Jaccard index 
	
The Jaccard index or Jaccard coefficient (J) is used to measure the spatial overlap of the intersection divided by the size of the union of two label sets. It can be expressed as shown in Equation (2), and can be obtained from the Dice measure by Equation (3).

	\begin{equation}
	J = \frac{\left|S\cap GT\right|}{\left|S \cup GT\right|}
	\end{equation}
	\begin{equation}
	J = \frac{DC}{(2-DC)}
	\end{equation}
		\item Volumetric Similarity (VS)
		
		VS is defined as the absolute volume difference divided by the sum of the compared volumes that are obtained by Equation (4). 
		
		\begin{equation}
		VS = 1 - {|\frac{\left|S\right|-\left|GT\right|}{\left|S\right| + \left| GT\right|}|}
		\end{equation}
		\begin{table}
		\centering
		\caption{\label{tab:similarity}Similarity measurements of the four different segmentation algorithms}
		\begin{tabular}{ |p{2cm}|p{1.5cm}|p{1.5cm}|p{1.5cm}|p{1.5cm}|p{1.5cm}|p{1.5cm}|}
			\hline
			\multicolumn{1}{|c|}{} & \multicolumn{2}{c}{DC}  & \multicolumn{2}{|c|}{Jaccard Index}  & \multicolumn{2}{c|}{Volumetric Similarity} \\ \hline
			
			Method & Mean &SD &Mean &SD	&Mean &SD \\ \hline
			Our Method & 0.942861 & 0.032463 & 0.933569 & 0.057144 & 0.967046 & 0.026951 \\ \hline
			2-Class CNN & 0.869014 & 0.048805 & 0.771535 & 0.077090 & 0.926983 & 0.047280 \\ \hline
			U-Net & 0.959566 & 0.000408 & 0.956901 & 0.000164 & 0.973044 & 0.002919 \\ \hline
			Level Set & 0.802039 & 0.068426 & 0.869837 & 0.012317 & 0.909029 & 0.089093 \\ \hline
		\end{tabular}
	\end{table}
\end{enumerate}
As demonstrated by Table \ref{tab:similarity}, our method shows a greater improvement in the segmentation results compared to other methods. Based on the DC score, we obtain ~94\%–97\% similarity with the ground truth. This result is also supported by the improvement in the volumetric similarity, which is ~96\% –98\%.

\item Classic measurements

We utilize the confusion matrix to perform classic measurements by utilizing four variables: true positive (TP), false positive (FP), true negative (TN), and false negative (FN).
\begin{itemize}
	\item TP: pixels correctly segmented as the spine in the ground truth and algorithm
	\item FP: pixels not classified as the spine in the ground truth, but are classified as the spine by algorithm (falsely segmented)
	\item TN:  pixels not classified as the spine in the ground truth and by algorithm (correctly detected as background)
	\item FN:  pixels classified as the spine in the ground truth, but not classified as the spine by the algorithm (falsely detected as background)
\end{itemize}
\begin{enumerate}[(i)]
	\item Sensitivity
	
Sensitivity measures the portion of positive pixels in the ground truth that are also identified as positive by the algorithm being evaluated. It is used to check algorithm sensitivity for detecting proper spine pixels. Sensitivity can be obtained by Equation (5)
	\begin{equation}
	Sensitivity = \frac{TP} {TP + FN}
	\end{equation}
	\item Specificity
	
Specificity measures the portion of negative pixels in the ground truth that are also identified as negative by the algorithm being evaluated. It checks how sensitive the algorithm is to the detection of correct background pixels. This metric is obtained by Equation (6)
	\begin{equation}
	Specificity = \frac{TN} {TN + FP}
	\end{equation}
	\item Over segmentation (OS) and under segmentation (US)
	
To further characterize the segmentation result, OS and US are obtained using Equations (7) and (8).. $GT^{-}$ and $S^-$ are the compliments of the gold standard and the segmentation results respectively.
	
	\begin{equation}
	OS =  \frac{2\left|S\cap GT^-\right|}{\left|S\right| + \left|GT\right|}\times 100
	\end{equation}
	\begin{equation}
	US =  \frac{2\left|S^-\cap GT\right|}{\left|S\right| + \left|GT\right|}\times 100
	\end{equation}
	\item Accuracy
	
	Accuracy is defined by Equation (9).
	\begin{equation}
	Accuracy = \frac{TP+TN}{TP+TN+FP+FN}
	\end{equation}
	\item Matthew correlations coefficient (MCC)
	
	MCC, introduced by B.W. Matthews, gives a summary of the performance of the segmentation algorithm. The MCC analyzes the segmentation result and the ground truth as two sets and takes into account TP and FN to compute a correlation coefficient that ranges between -1 (complete disagreement) and 1 (complete agreement). A value of zero shows that the segmentation was not correlated with the ground truth. The MCC can be defined as shown in Equation (10)
	\begin{equation}
	MCC = \frac{TP\times TN - FP\times FN}{\sqrt{(TP + FN)\times (TP + FP)\times (TN + FP)\times (TN + FN)}}
	\end{equation}
	
	\begin{table}
		\centering
			\caption{\label{tab:classic}Classic measurements of the four different segmentation algorithms}
		\begin{tabular}{ |p{2cm}|p{1.5cm}|p{1.5cm}|p{1.5cm}|p{1.5cm}|p{1.5cm}|p{1.5cm}|}
			\hline
			\multicolumn{1}{|c|}{} & \multicolumn{2}{c}{Sensitivity}  & \multicolumn{2}{|c|}{Specificity}  & \multicolumn{2}{c|}{Over Segmentation} \\ \hline
			
			Method & Mean &SD &Mean &SD	&Mean &SD \\ \hline
			Our Method & 0.972481& 0.026948 & 0.991616 & 0.006239 & 8.349675 & 5.186457 \\ \hline
			2-Class CNN & 0.913395 & 0.077838 & 0.973564 & 0.019650 & 16.26140 & 8.085861 \\ \hline
			U-Net & 0.972806 & 0.003227 & 0.998899 & 0.000166 & 1.356735 & 0.248874\\ \hline
			Level Set & 0.779723 & 0.123521 & 0.987453 & 0.009751 & 15.15807 & 9.973117 \\ \hline
		\end{tabular}
	
	\end{table}
	
\begin{table}
	\centering
	\begin{tabular}{ |p{2cm}|p{1.5cm}|p{1.5cm}|p{1.5cm}|p{1.5cm}|p{1.5cm}|p{1.5cm}|}
		\hline
		\multicolumn{1}{|c|}{} & \multicolumn{2}{c}{Under Segmentation}  & \multicolumn{2}{|c|}{Accuracy}  & \multicolumn{2}{c|}{MCC} \\ \hline
		
		Method & Mean &SD &Mean &SD	&Mean &SD \\ \hline
		Our Method & 2.643167 & 2.667397 & 99.01979 & 0.575767 & 0.938438 & 0.033661 \\ \hline
		2-Class CNN & 8.944368 & 9.749133 & 96.76865 & 1.907359 & 0.852528 & 0.047596 \\ \hline
		U-Net & 2.758809 & 0.343256 & 99.69015 & 0.016499 & 0.977921 & 0.000479 \\ \hline
		Level Set & 27.792356 & 22.945279 & 96. 98377 & 1.231796 & 0.792400 & 0.066891 \\ \hline
	\end{tabular}
\end{table}
\end{enumerate}
Our method shows the greatest improvement in the sensitivity and specificity categories. Sensitivity is ~97\%, specificity is ~99\%, and the accuracy is ~99\%, which is similar to the U-Net method. The results also showed a significant improvement in the OS and US categories compared to the classic CNN method. These results are also supported by the improvement in the MCC, which showed that our segmentation results achieved very good agreement with the ground truth.
\item Distance measurements
\begin{enumerate}[(i)]
	\item Mean surface distance (MSD)
	
	MSD is the mean of the absolute values of the surface distance for all $n_s$ surface voxels. This value is obtained by Equation (11).
	\begin{equation}
	MSD_{sg} = \frac{1}{n_s} \sum_{i = 1}^{n_s} \left|d_l^{sg}\right|
	\end{equation}
	This metrics attempts to estimate the error between the surfaces of S and GT using distances between their surface voxels. We define the surface distance of the $i^{th}$ surface voxel on S, $d_l^{sg}$ as the distance to the closest voxel on GT. The surface distance values were calculated as described in Gerig et al. \cite{ref31}. In the distance transform image, the value of each voxel is the Euclidean distance in millimeters to the nearest surface voxel. The values on the surface voxels are 0.

	\item Hausdorff distance (HD)
	
	HD measures the distance between the ground truth surface and the segmented surface. To compute the HD, we use the same surface models that were generated to compute the MSD. A smaller HD indicates better segmentation accuracy. This metric can be defined by Equation (12)
	\begin{equation}
	HD = \max(h(S,GT),h(GT,S))
	\end{equation}
	where, $\displaystyle{h(S,GT)} = \max_{a\in S}\min_{b\in GT}\norm{a-b}$
	\item Global consistency Error (GCE)
	
	GCE is defined as an error measurement between two segmentations (the error averaged over all voxels) and is given by Equation (13)
	
	\begin{equation}
	GCE(S,GT) = \frac{1}{N} \min{\left\lbrace\sum_{i}^{n}E(S,GT,x_i),\sum_{i}^{n}E(GT,S,x_i)\right\rbrace}
	\end{equation}
	where, $N$ is the total voxel, $n$ is the set difference and the error $E$ at voxel $x$ is defined as $E(S,GT,x) = \frac{\left|(S,x) n (GT,x)\right|}{(S,x)}$ 
	
\end{enumerate}
    \begin{table}
	\centering
	\caption{\label{tab:distance}Distance measurements of the four different segmentation algorithms}
	\begin{tabular}{ |p{2cm}|p{1.5cm}|p{1.5cm}|p{1.5cm}|p{1.5cm}|p{1.5cm}|p{1.5cm}|}
		\hline
		\multicolumn{1}{|c|}{} & \multicolumn{2}{c}{MSD}  & \multicolumn{2}{|c|}{HD}  & \multicolumn{2}{c|}{GCE} \\ \hline
		
		Method & Mean &SD &Mean &SD	&Mean &SD \\ \hline
		Our Method & 0.167692 & 0.229274 & 7.429119 & 1.635649 & 0.020012 & 0.010673 \\ \hline
		2-Class CNN & 0.784873 & 0.437554 & 9.502033 & 1.803488 & 0.064253 & 0.039871 \\ \hline
		U-Net & 0.099893 & 0.005112 & 6.289700 & 0.491765 & 0.030092 & 0.001779\\ \hline
		Level Set & 0.342764 & 0.352889 & 8.788802 & 4.781518 & 0.054126 & 0.021791 \\ \hline
	\end{tabular}
\end{table}
Based on the distance metrics, our methods obtained better results than the classic CNN and the level set methods. We showed improvement and achieved good results based on the GCE and both distance metrics. The GCE distribution of the proposed method is close to zero, which indicates a very low error segmentation result.

\end{enumerate}
\section{Conclusion and discussion}
In this study, a new approach for medical image segmentation was proposed. This approach uses class redundancy as a soft constraint in the CNN architecture. The proposed method achieved a competitive result compared to several widely used medical image segmentation methods. The proposed algorithm was tested on real medical data and evaluated through similarity metrics such as the Dice coefficient, Jaccard index, volumetric similarity, sensitivity, specificity, precision (over segmentation and under segmentation), accuracy, Matthews correlation coefficient, mean surface distance, Hausdorff distance, and global consistency error. These similarity metrics generated results of 94\%, 93\%, 96\%, 97\%, 99\%, 8.3, 2.6, 99\%, 0.93, 0.16, 7.4, and 0.02, respectively. These results demonstrate the effectiveness of our method. 

In conclusion, the main contribution of this study is the presentation of a new approach to prepare data and its use in a simple CNN model to improve the accuracy of segmentation results. The experimental results quantitatively and qualitatively showed that our proposed method improves accuracy, corrects error segmentation, and exhibits better segmentation performance compared to other conventional methods. The results also showed high specificity and sensitivity, and high overlap and low distance between the manual annotation and the proposed method.

In future, we intend to assess our method using a deeper network and broad training data, such as low contrast data, to improve the performance. The hope is to eventually be capable of handling a broader range of medical data. Another possible direction is to investigate improvements to computation time during the training stage through optimization. Currently, for 1.388.800 patches of spine data, it takes ~13 h to train using our CPU, and processing time takes ~3 min for each image. 
\section{Acknowledgment}
This research was supported by WC300 Project (2E26210) funded by the Ministry of SMEs and Startups (MSS, Korea) [Project NO. S2482672].

\nocite{ref1,ref2,ref3,ref4,ref5,ref6,ref8,ref9,ref10,ref11,ref12,ref13,ref14,ref15,ref16,ref17,ref18,ref19,ref20,ref21,ref22,ref23,ref24,ref25,ref26,ref27,ref28,ref29,ref30,ref31}
\section*{References}

\bibliography{mybibfile}

\begin{thebibliography}{10}
\expandafter\ifx\csname url\endcsname\relax
  \def\url#1{\texttt{#1}}\fi
\expandafter\ifx\csname urlprefix\endcsname\relax\def\urlprefix{URL }\fi
\expandafter\ifx\csname href\endcsname\relax
  \def\href#1#2{#2} \def\path#1{#1}\fi

\bibitem{ref3}
G.~Kumar, Analysis of medical image processing and its applications in
  healthcare industry, International Journal of Computer Technology \&
  Applications 5~(3) (2014) 851--860.

\bibitem{ref2}
M.~Shoham, I.~Lieberman, E.~Benzel, E.~Zehavi, B.~Zilberstein, M.~Roffman,
  A.~Bruskin, A.~Fridlander, L.~Joskowicz, S.~Brink-Danan, N.~Knoller, Robotic
  assisted spinal surgery: from concept to clinical practice, Computer Aided
  Surgery, 12 (2007) 105--115.

\bibitem{ref1}
D.~L. Pham, C.~Xu, J.~Prince, Current methods in medical image segmentation,
  Annual Review of Biomedical Engineering, 2 (2000) 315--337.

\bibitem{ref4}
Elsevier, I.~Bankman, Handbook of medical image processing and analysis, 2nd
  ed. (2009) 71--257.

\bibitem{ref5}
G.~Harris, N.~C. Andreasen, J.~M.~B. T.~Cizadlo, H.~J. Bockholt, V.~A.
  Magnotta, S.~Arndt, Improving tissue classification in {MRI}: A
  three-dimensional multispectral discriminant analysis method with automated
  training class selection, Journal of Computer Assisted Tomography, 23~(1)
  (1999) 144--154.

\bibitem{ref6}
T.~Klinder, J.~Ostermann, M.~Ehm, A.~Franz, R.~Kneser, C.~Lorenz, Automated
  model-based vertebra detection identification and segmentation in ct images,
  Medical Image Analysis, 13~(3) (2009) 471--482.

\bibitem{ref8}
M.~Krcah, G.~Szekely, R.~Blanc, Fully automatic and fast segmentation of the
  femur bone from 3{D}--{CT} images with no shape prior, Proceedings of IEEE
  International Symposium of Biomedical Imaging, (2011) 2087--2090.

\bibitem{ref11}
M.~Hardisty, L.~Gordon, P.~Agarwal, T.~Skrinskas, C.~Whyne, Quantitative
  characterization of metastatic disease in the spine. {Part I}. {Semiautomated
  segmentation using atlas-based deformable registration and the level set
  method}, Medical Physics, 34~(8) (2007) 3127--3134.

\bibitem{ref12}
L.~Gordon, M.~Hardisty, T.~Skrinskas, F.~Wu, C.~Whyne, Automated atlas-based
  {3D} segmentation of the metastatic spine, Journal of Bone \& Joint Surgery,
  90 (2008) 128.

\bibitem{ref13}
Y.~Kang, K.~Engelke, W.~A. Kalender, A new accurate and precise {3D}
  segmentation method for skeletal structures in volumetric {CT} data, IEEE
  Transactions on Medical Imaging, 22 (2003) 586--598.

\bibitem{ref14}
A.~Mastmeyer, K.~Engelke, C.~Fuchs, W.~A. Kalender, A hierarchical {3D}
  segmentation method and the definition of vertebral body coordinate systems
  for {QCT} of the lumbar spine, Medical Image Analysis, 10 (2006) 560--577.

\bibitem{ref15}
G.~Sambuceti, M.~Brignone, a.~M.~M. Marini, C., F.~Fiz, S.~Morbelli,
  A.~Buschiazzo, C.~Campi, R.~Piva, A.~Massone, M.~Piana, F.~Frassoni,
  Estimating the whole bone-marrow asset in humans by a computational approach
  to integrated {PET}/{CT} imaging, European Journal of Nuclear Medicine,
  Molecular Imaging, 39~(8) (2012) 1326--1338.

\bibitem{ref16}
J.~Yao, S.~O’Connor, R.~Summers, Automated spinal column extraction and
  partitioning, Proceedings of IEEE International Symposium of Biomedical
  Imaging, (2006) 390--393.

\bibitem{ref17}
T.~Klinder, J.~Ostermann, M.~Ehm, A.~Franz, R.~Kneser, C.~Lorenz, Automated
  model-based vertebra detection, identification, and segmentation in {CT}
  images, Medical Image Analysis, 13~(3) (2009) 471--482.

\bibitem{ref18}
S.~Huang, Y.~Chu, S.~Lai, C.~Novak, Learning-based vertebra detection and
  iterative normalized-cut segmentation for spinal {MRI}, {IEEE} Transactions
  on Medical Imaging, 28~(10) (2009) 1596--1605.

\bibitem{ref19}
J.~Ma, L.~Lu, Hierarchical segmentation and identification of thoracic vertebra
  using learning-based edge detection and coarse\-to\-fine deformable model,
  Computer Vision and Image Understanding, 117~(9) (2013) 1072--1083.

\bibitem{ref20}
B.~Glocker, D.~Zikic, E.~Konukoglu, D.~Haynor, A.~Criminisi, Vertebrae
  localization in pathological spine {CT} via dense classification from sparse
  annotations, Proceedings of International Conference on Medical Computing and
  Computer Assisted Intervention, (2013) 262--270.

\bibitem{ref21}
S.~Kim, M.~Kang, Multiple-region segmentation without supervision by adaptive
  global maximum clustering, IEEE Transactions on Image Processing, 21~(4)
  (2012) 1600--1612.

\bibitem{ref22}
S.~Kim, Y.~Kim, S.~Park, D.~Lee, Automatic segmentation of leg bones by using
  active contours, Conf Proc IEEE Eng Med Biol Soc., (2014) 4695--4698.

\bibitem{ref23}
M.~Vania, S.~Kim, D.~Lee, Automatic multisegmentation of abdominal organs by
  level set with weighted global and local forces, Student Conference (ISC),
  2016 IEEE EMBS International.

\bibitem{ref28}
Y.~LeCun, B.~Boser, J.~S. Denker, D.~Henderson, R.~E. Howard, W.~Hubbard, L.~D.
  Jackel, Backpropagation applied to handwritten zip code recognition, Neural
  Computation, 1~(4) (1989) 541--551.

\bibitem{ref29}
A.~Krizhevsky, S.~Ilya, G.~E. Hinton, Imagenet classification with deep
  convolutional neural networks, Advances in Neural Information Processing
  Systems 25 (2012) 1097--1105.

\bibitem{ref26}
J.~Long, E.~Shelhamer, T.~Darrell, Fully convolutional networks for semantic
  segmentation, Proceedings of the IEEE Conference on Computer Vision and
  Pattern Recognition, (2015) 3431--3440.

\bibitem{ref27}
O.~Ronneberger, P.~Fischer, T.~Brox, U-net: Convolutional networks for
  biomedical image segmentation, Medical Image Computing and Computer-Assisted
  Intervention MICCAI 2015, (2015) 234--241.

\bibitem{ref24}
A.~A. Taha, A.~Hanbury, Metrics for evaluating {3D} medical image segmentation:
  analysis, selection, and tool. bmc medical imaging, {BMC} Medical Imaging
  (2015) 15--29.

\bibitem{ref25}
J.~Udupa, V.~Leblanc, Y.~Zhuge, C.~Imielinska, H.~Schmidt, L.~Currie,
  B.~Hirsch, J.~Woodburn, A framework for evaluating image segmentation
  algorithms, Computerized Medical Imaging and Graphics, 30~(2) (2006) 75--87.

\bibitem{ref31}
G.~Gerig, M.~Jomier, M.~Chakos, Valmet: a new validation tool for assessing and
  improving 3d object segmentation, Proceedings of the 4th International
  Conference on Medical Image Computing and Computer-Assisted Intervention
  (2001) 516--523.

\bibitem{ref9}
S.~Lashari, R.~A. Ibrahim, A framework for medical images classification using
  soft set, Procedia Technology, 11 (2013) 548--556.

\bibitem{ref10}
D.~Yang, J.~Zheng, A.~Nofal, J.~Deasy, I.~M.~E. Naqa, Techniques and software
  tool for {3D} multimodality medical image segmentation, Journal of Radiation
  Oncology Informatics, 1~(1).

\bibitem{ref30}
I.~Goodfellow, Y.~Bengio, A.~Courville, Deep learning, MIT press (2016)
  321--363.

\end{thebibliography}

\end{document}